%% file: main.tex
\documentclass[10pt,twocolumn,letterpaper]{article}
\PassOptionsToPackage{dvipsnames,table,xcdraw}{xcolor}
\usepackage{iccv}              


\definecolor{iccvblue}{rgb}{0.21,0.49,0.74}
\usepackage[pagebackref,breaklinks,colorlinks,allcolors=iccvblue]{hyperref}
\usepackage{algorithm}
\usepackage{algorithmicx}
\usepackage{algpseudocode}
\usepackage{amsmath}

\usepackage{makecell}
\usepackage{booktabs}
\usepackage{multirow}
\usepackage{graphicx}
\usepackage{array}
\usepackage{adjustbox}
\usepackage{colortbl}
\usepackage{tabularx}
\definecolor{lightgray}{gray}{0.9}
\definecolor{deepgreen}{HTML}{54B345}
\newcommand{\tablestyle}[2]{\setlength{\tabcolsep}{#1}\renewcommand{\arraystretch}{#2}\centering\small}
\newcommand{\hgreen}[1]{\textcolor{ForestGreen}{\textbf{#1}}} 

\def\paperID{2723} 
\def\confName{ICCV}
\def\confYear{2025}

\title{VFlowOpt: A Token Pruning Framework for LMMs with Visual Information Flow-Guided Optimization}

\author{
  Sihan Yang$^{1}$ \quad
  Runsen Xu$^{1,2}$ \quad
  Chenhang Cui$^{3}$ \quad
  Tai Wang$^{1\dagger}$ \quad
  Dahua Lin$^{1,2}$ \quad
  Jiangmiao Pang$^{1\dagger}$ \\
  \small
  $^{1}$ Shanghai AI Laboratory \quad
  $^{2}$ The Chinese University of Hong Kong \quad
  $^{3}$ National University of Singapore \\
  $^{\dagger}$~Corresponding Author \\
  {\tt taiwang.me@gmail.com \quad pangjiangmiao@gmail.com}
}
\begin{document}
\maketitle

\input{sec/0_abstract}    
\input{sec/1_intro}
\input{sec/2_related_work}

\input{sec/3_method}

\input{sec/4_experiments}
\input{sec/5_conclusion}

\section*{Acknowledgements}
This research was supported by Shanghai Artificial Intelligence Laboratory.

{
    \small
    \bibliographystyle{ieeenat_fullname}
    \bibliography{main}
}

\clearpage
\newpage 
\input{sec/suppl}

\end{document}

%% file: sec/0_abstract.tex
\begin{abstract}
Large Multimodal Models (LMMs) excel in visual-language tasks by leveraging numerous visual tokens for fine-grained visual information, but this token redundancy results in significant computational costs. Previous research aimed at reducing visual tokens during inference typically leverages importance maps derived from attention scores among vision-only tokens or vision-language tokens to prune tokens across one or multiple pruning stages. Despite this progress, pruning frameworks and strategies remain simplistic and insufficiently explored, often resulting in substantial performance degradation. In this paper, we propose VFlowOpt, a token pruning framework that introduces an importance map derivation process and a progressive pruning module with a recycling mechanism. The hyperparameters of its pruning strategy are further optimized by a visual information flow-guided method. Specifically, we compute an importance map for image tokens based on their attention-derived context relevance and patch-level information entropy. We then decide which tokens to retain or prune and aggregate the pruned ones as recycled tokens to avoid potential information loss. Finally, we apply a visual information flow-guided method that regards the last token in the LMM as the most representative signal of text-visual interactions. This method minimizes the discrepancy between token representations in LMMs with and without pruning, thereby enabling superior pruning strategies tailored to different LMMs. Experiments demonstrate that VFlowOpt can prune 90\% of visual tokens while maintaining comparable performance, leading to an 89\% reduction in KV-Cache memory and 3.8$\times$ faster inference. Code is at \href{https://github.com/sihany077/VFlowOpt}{https://github.com/sihany077/VFlowOpt}.\end{abstract}

%% file: sec/1_intro.tex
\section{Introduction}
Large Multimodal Models (LMMs)~\cite{liu2023improvedllava,liu2023llava,Qwen-VL,chen2024internvl} have attained remarkable performance in tasks such as visual question answering~\cite{Antol_2015_ICCV,goyal2017makingvvqamatter} and multimodal reasoning~\cite{yue2023mmmu,yue2024mmmuprorobustmultidisciplinemultimodal}, making them indispensable for applications like autonomous driving~\cite{cui2023surveymultimodallargelanguage,wang2023drivemlmaligningmultimodallarge,tian2024llm} and robotics~\cite{li2023manipllmembodiedmultimodallarge,liu2024selfcorrectedmultimodallargelanguage,li2025integrating}. To capture fine-grained visual details, modern LMMs treat images as token sequences, with models such as LLaVA-1.5~\cite{liu2023improvedllava} processing hundreds of tokens and LLaVA-OneVision~\cite{li2024llavaonevisioneasyvisualtask} handling up to several thousand tokens. Nonetheless, the increasing quantity of visual tokens significantly magnifies computational cost, memory usage, and inference time, creating a critical bottleneck for LMM deployment, particularly in resource-constrained environments or latency-sensitive scenarios.
 
\begin{figure}[t]
    \centering
    \vspace{-0.3cm}
    \includegraphics[width=1\linewidth]{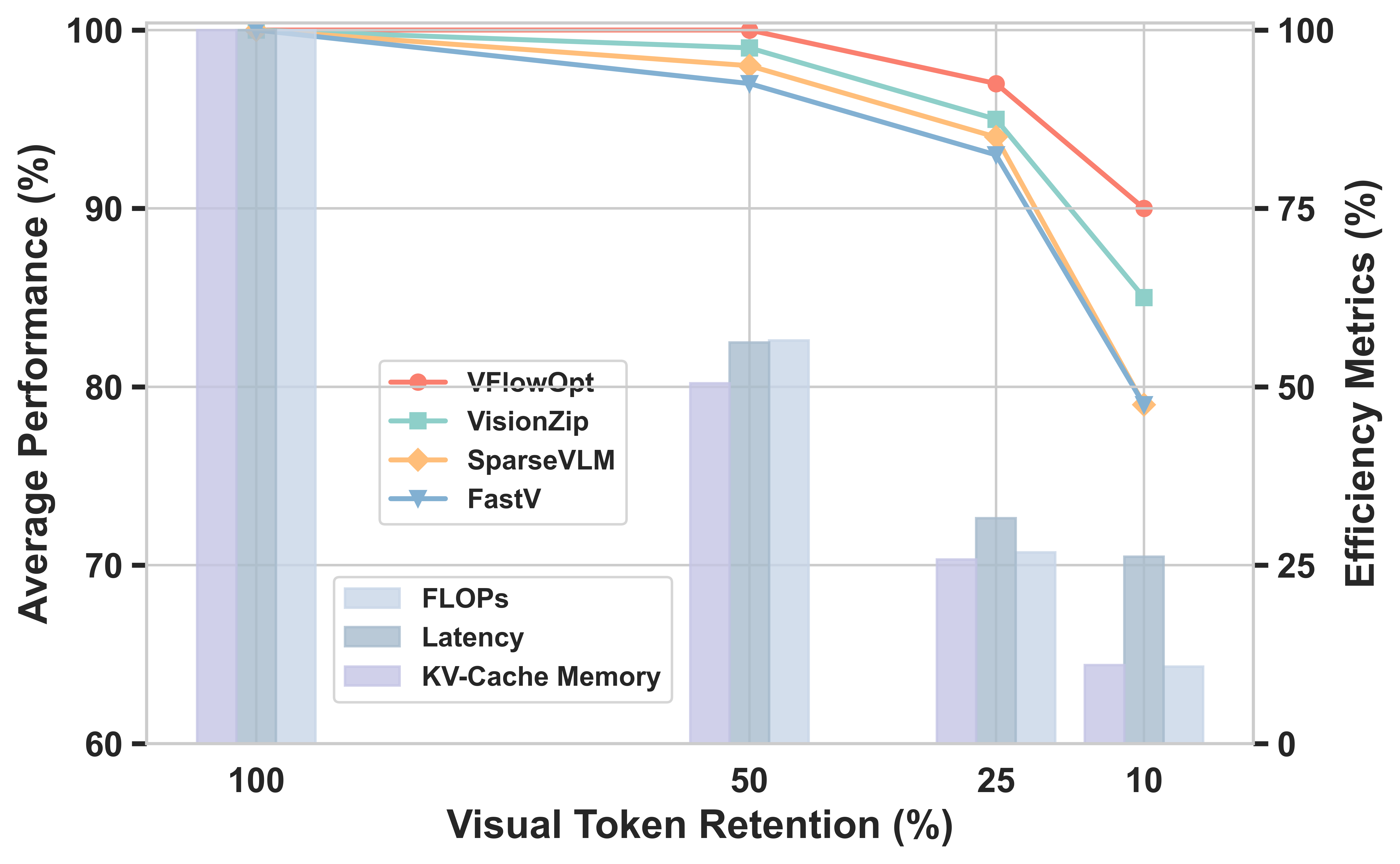}
        \vspace{-0.7cm}
    \caption{
    \textbf{VFlowOpt Performance and Efficiency.} Our VFlowOpt significantly outperforms other LMM token reduction methods across three general benchmarks (MME, MMStar, and MMBench) on LLaVA-OneVision-7B. It achieves zero performance degradation when reducing 50\% of visual tokens. Furthermore, when retaining only 10\% of the tokens, it achieves 90\% of the original performance while reducing KV-Cache memory usage by 89\% and accelerating inference by 3.8$\times$.
    }
    \vspace{-0.5cm}
    \label{fig:teaser}
\end{figure}
To alleviate these constraints, recent work explores various methods for reducing the number of visual tokens during inference. Most existing approaches evaluate the importance of visual tokens in LMMs using attention-based importance maps derived either from vision-only tokens~\citep{Visionzip,zhang2024clsattentionneedtrainingfree} or from text-visual interactions~\citep{Sparsevlm,Pdrop}. Following a predefined pruning ratio, visual tokens are pruned in a single stage~\cite{wang2024cls} or progressively across multiple stages~\cite{Fitprune}, often with heuristic strategies applied uniformly across different LMMs. Although promising, these methods are simplistic and underexplored, with coarse-grained pruning often causing significant performance drops due to lacking fitness for different LMMs models' characteristics.

In response to these limitations, we propose a novel token pruning framework, VFlowOpt, that more accurately estimates token importance by integrating attention calibration and entropy, and reduces information loss through token recycling and progressive pruning strategy with fine-grained pruning ratios. Furthermore, by customizing pruning strategies for different models, our framework better preserves model performance.
Specifically, the pruning framework consists of importance map computation and a progressive pruning module with a recycling mechanism. We decompose the importance map of image tokens into two aspects: the importance of tokens in the visual context, reflected by their attention maps, and the richness of visual information in each image patch, captured by its information entropy. Based on the weighted summation of these two factors, we determine which tokens to retain or prune across multiple pruning stages and introduce a recycling mechanism to aggregate pruned tokens to avoid potential information loss. Built with this pruning strategy with several hyperparameters, a key ingredient is the visual information flow-guided method to optimize the pruning strategy, which treats the last token in each pruning stage as the most representative signal of text-visual interactions during inference. By minimizing the difference between token representations in LMMs with and without pruning, VFlowOpt ultimately delivers superior pruning strategies specifically tailored to different LMMs.

Comprehensive experiments on multiple vision-language benchmarks validate the efficacy of VFlowOpt. In particular, it substantially lowers computational cost while preserving competitive performance. As shown in ~\cref{fig:teaser}, VFlowOpt can prune 50\% of visual tokens with negligible performance loss. Moreover, it can prune 90\% of visual tokens while maintaining 90\% of the model's performance, resulting in an 89\% reduction in KV-Cache memory usage and a 3.8$\times$ speedup in inference. These experimental results highlight its effectiveness as a practical and efficient solution for deploying LMMs in resource-constrained environments.

%% file: sec/2_related_work.tex
\section{Related Work}
\subsection{Large Multimodal Models}
Large Multimodal Models (LMMs)~\cite{liu2023improvedllava,liu2023llava,Qwen-VL,chen2024internvl} extend the reasoning capabilities of Large Language Models (LLMs)~\cite{touvron2023llama2openfoundation,qwen2,qwen2.5,cai2024internlm2} to vision-language tasks by integrating a pre-trained vision encoder~\cite{radford2021learning,zhai2023sigmoid} with a language model, linked by an alignment module such as an MLP, or a query-based network. This design transforms visual inputs into token sequences that the LLM can process, facilitating multimodal prompts for tasks like visual question answering~\cite{Antol_2015_ICCV,goyal2017makingvvqamatter,MMstar,yue2023mmmu}. To enhance performance, advanced LMMs, such as LLaVA-OneVision~\cite{li2024llavaonevisioneasyvisualtask} and Qwen2-VL~\cite{wang2024qwen2}, can encode higher-resolution images into more image tokens, thereby capturing more granular visual details. However, as image resolution grows, the number of visual tokens rises exponentially, leading to substantially higher computational costs. For example, LLaVA-1.5~\cite{liu2023improvedllava} processes 336×336 images into 576 tokens, whereas LLaVA-OneVision can handle 1152×1152 images, producing 7,290 tokens. This challenge becomes even more pronounced in video-based models like LongVA~\cite{zhang2024long}, which must process tokens across numerous frames. Fine-grained visual tokenization boosts LMM performance but poses an inference bottleneck, driving efforts to balance performance and cost through token reduction.

\subsection{Token Reduction for LMMs}
Token reduction~\cite{chai2024auroracap,li2025catp} has become a key strategy for improving the efficiency of LMMs by mitigating the computational cost associated with extensive visual token sequences. Existing methods can be broadly classified into training-based and training-free approaches. Training-based methods, such as LLaVA-Mini~\cite{zhangllava} and LLaVolta~\cite{chen2024llavolta}, introduce additional modules during model training to compress visual tokens and preserve critical information, while approaches like ATP-LLaVA~\cite{ye2024atp} and p-MoD~\cite{zhang2024p} train pruning modules to dynamically retain important tokens across LLM layers. However, these methods demand substantial computational resources to retrain the models, limiting their real-world applicability. In contrast, training-free methods prune tokens without additional training, often leveraging attention mechanisms to identify and discard redundant tokens. For instance, FastV\cite{FastV} and SparseVLM exploit text–visual attention to rank token importance, whereas FasterVLM~\cite{zhang2024clsattentionneedtrainingfree}, VisionZip~\cite{Visionzip}, and VTC-CLS~\cite{wang2024cls} rely on [CLS] token attention in the vision encoder to evaluate token importance and prune redundant tokens. FitPrune~\cite{Fitprune} and PDrop~\cite{Pdrop} propose a progressive multi-stage pruning strategy in LMMs to fully utilize visual information. However, the pruning strategies in these methods often lack adaptability to different LMMs, which frequently leads to significant performance drops. 

%% file: sec/3_method.tex
\begin{figure*}[t] 
  \centering
\includegraphics[width=0.99\linewidth]{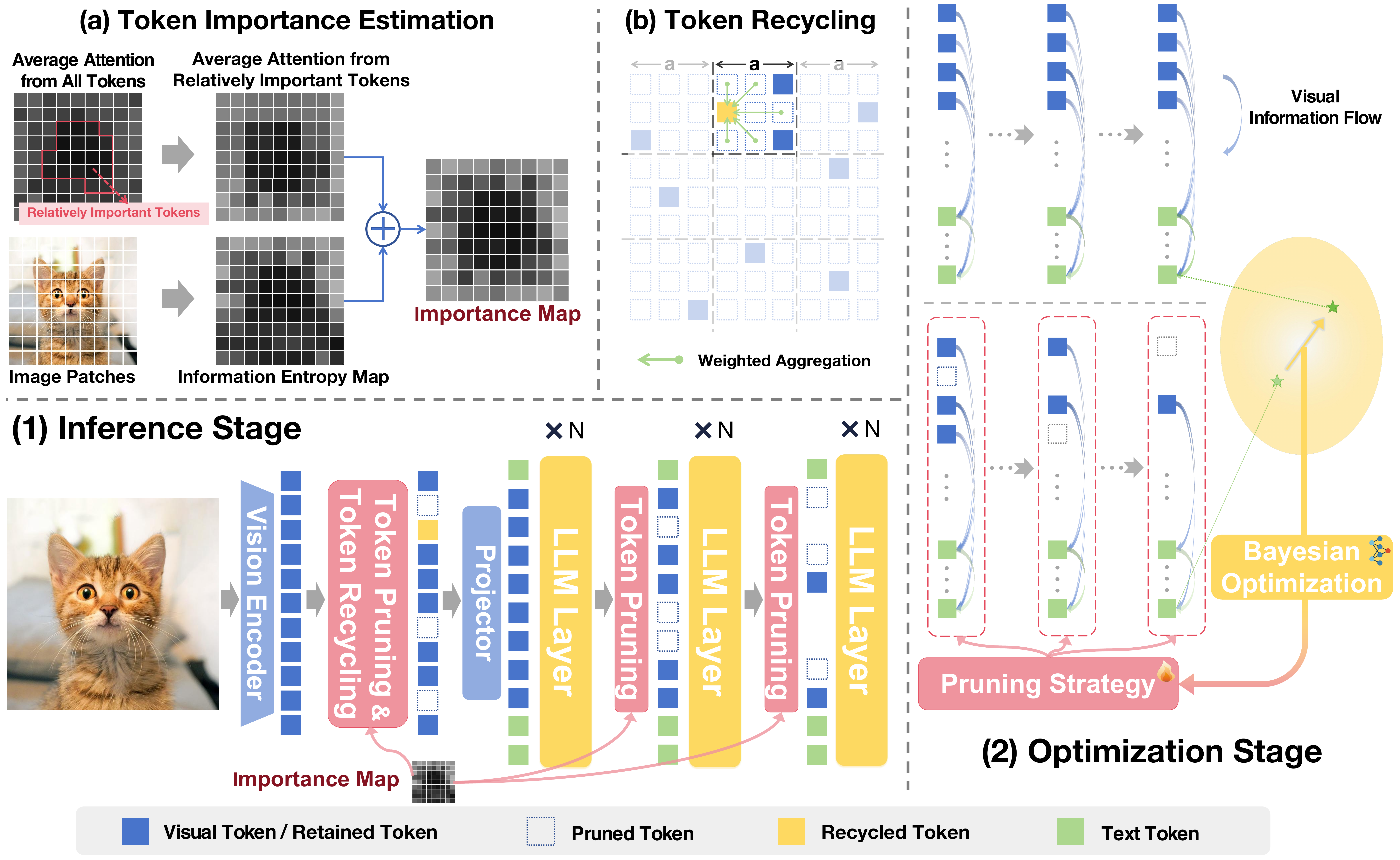} 
  \caption{\textbf{Overview of VFlowOpt.} (1) During inference, VFlowOpt first assesses the importance of visual tokens, based on which progressive token pruning is performed. After the initial pruning stage, the pruned tokens are merged and recycled. The pruning strategy used in this process is defined by the (2) Optimization Stage. (a) The importance map is computed by combining the attention of relatively important tokens with the entropy of image patches. (b) The pruned tokens are grouped into grid cells, where each cell has a side length of a. Within each grid cell, the pruned tokens are fused using a weighted average, with their importance values as weights, and then recycled. (2) VFlowOpt optimizes the pruning strategy by minimizing the discrepancy of the last token in the final layer of the LMM with and without applying the pruning strategy.}   
\label{fig:inference_framework} 
\end{figure*}
\begin{figure}[h!] 
  \centering
  \includegraphics[width=0.92\linewidth]{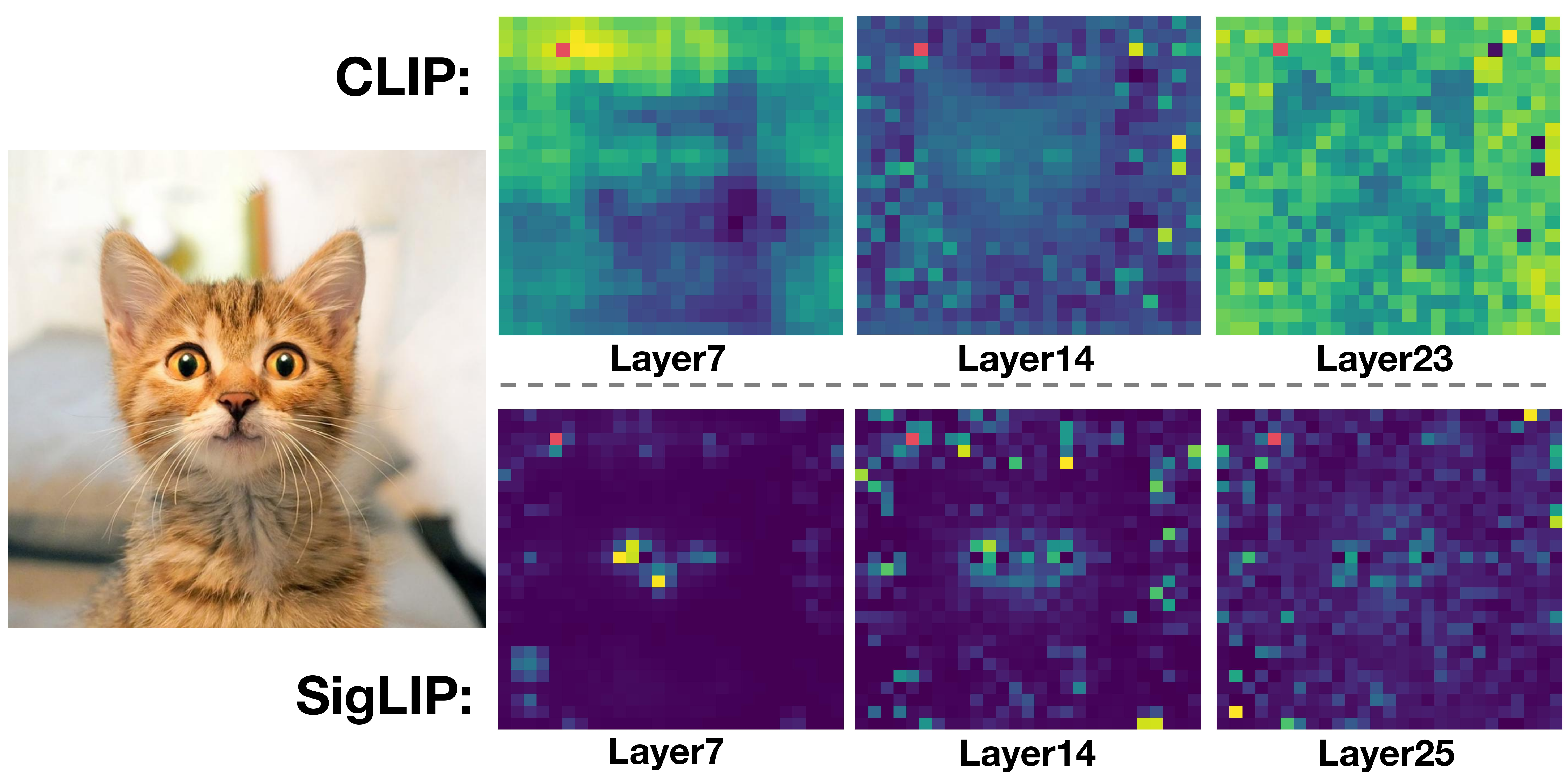} 
  \vspace{-0.1cm}
  \caption{The attention of redundant tokens (marked in red) fails to reflect the importance of other tokens and instead focuses on similar tokens, such as background elements.}
  \label{fig:reductant_token_attn} %
\end{figure}
\section{Method}
In this section, we introduce VFlowOpt, a framework that employs a pruning strategy to reduce redundant visual tokens during LMM inference while preserving essential visual information. This strategy involves multiple hyperparameters that our VFlowOpt framework automatically optimizes for proper configuration—an essential factor for maintaining model performance. Specifically, \Cref{sec:3_1} explains how our method evaluates the importance of visual tokens, \Cref{sec:3_2} presents progressive pruning and token recycling strategies, and finally, \Cref{sec:3_3} details how VFlowOpt customizes the superior pruning strategy (i.e., selects the pruning hyperparameters) for different LMMs.

\subsection{Visual Token Importance Estimation}
\label{sec:3_1}

 For a general approach to evaluate visual token importance, the previous works~\cite{yang2024visionzip,hu2024illava} propose that the importance of visual tokens can be estimated using the average attention from all other tokens in ViTs. Although it has achieved some success, we find that redundant tokens (e.g., tokens corresponding to background regions) often assign disproportionately high attention to other similar redundant tokens (shown in \cref{fig:reductant_token_attn}), reducing the reliability of the importance estimation. To address this, we first identify relatively important tokens based on the attention they receive from all tokens (tokens receiving higher attention are considered relatively important), as shown in \cref{fig:inference_framework} (a). We then exclude redundant tokens and use the attention from these relatively important tokens as a more robust metric. Additionally, we incorporate the information entropy of image patches into the importance score to prioritize tokens corresponding to visually informative regions.

To identify relatively important visual tokens, we define a threshold:
\begin{equation}
    \tau = t \cdot \frac{1}{N} \sum_{i=1}^{N} \sum_{j=1}^{N} A_{ij},
\end{equation}
where \( \tau \) is the threshold, \( A_{ij} \) is the attention weight from token \( i \) to token \( j \) within a ViT layer, \( N \) is the total number of tokens, and \( t \) is a sensitivity hyperparameter. We then treat tokens with a total attention exceeding \( \tau \) as relatively important:
\begin{equation}
    \mathcal{K} = \{ j \mid \sum_{i=1}^{N} A_{ij} > \tau \}.
\end{equation}
Here, \( \mathcal{K} \) denotes the set of indices of these relatively important tokens. 

The information entropy of an image patch corresponding to token \( i \) is defined as:
\begin{equation}
    H(V_i) = - \sum_{k=0}^{L-1} p_k \log p_k,
\end{equation}
where \( V_i \) is the image patch corresponding to token \( i \), \( p_k \) is the proportion of pixels in the patch with gray level \( k \), and \( L \) is the total number of possible gray levels (256 for 8-bit images). Here, the gray level is defined as the average of the RGB channel values, providing a scalar representation of pixel brightness. Higher entropy reflects greater diversity in pixel intensities and richer visual information.

The importance score of token \( i \) is computed by combining the attention it receives from tokens in \(\mathcal{K}\) with the entropy of its corresponding image patch, normalized via the softmax function:
\begin{equation}
    I_i = \sum_{k \in \mathcal{K}} A_{ki} \;+\; \alpha \cdot \frac{\exp\bigl(H(V_i)\bigr)}{\sum_{j=1}^{N} \exp\bigl(H(V_j)\bigr)},
\end{equation}
where \( I_i \) denotes the importance score of token \( i \), \( A_{ki} \) is the attention weight from token \(\ k \) to token \( i \), \( H(V_i) \) is the entropy of the image patch corresponding to token \( i \), \(\alpha\) is a hyperparameter that determines the entropy term’s contribution, and \( N \) is the total number of tokens. This approach mitigates biases from redundant tokens and leverages the visual information of image patches, thereby facilitating effective estimation of visual token importance.

\subsection{Progressive Pruning and Token Recycling}
\label{sec:3_2}
Previous studies~\cite{zhang2024redundancy,xing2024pyramiddrop} indicate that visual tokens play a more critical role in the shallower layers of LMMs, whereas redundancy tends to increase in deeper layers. Based on this, we adopt a progressive pruning strategy. To balance the pruning process's simplicity with its fine-grained configuration, we evenly divide the LMM into three stages. At the beginning of each stage, a predetermined fraction of visual tokens with higher importance scores (as defined in \cref{sec:3_1}) is retained according to the stage-specific retention ratios \( R = [R_1, R_2, R_3] \), while the rest are pruned. Notably, the position IDs of visual tokens remain unchanged following pruning, preserving the original spatial structure of the visual input. Following the previous work~\cite{xing2024pyramiddrop}, we compute the average visual token retention rate across the entire LMM as follows:
\begin{equation}
\overline{R} = \frac{R_1 \cdot L_1 + R_1 \cdot R_2 \cdot L_2 + R_1 \cdot R_2 \cdot R_3 \cdot L_3}{L},
\end{equation}
where \(L_1\), \(L_2\), and \(L_3\) represent the number of layers in the three stages, respectively, and \(L\) denotes the total number of layers in the LMM.

To prevent the loss of any small but potentially significant information during the initial pruning process (before visual tokens are fed into the LLM), we propose a token merging and recycling strategy to compactly represent redundant visual information. Specifically, We use a square grid (with each cell having side length \(a\)) to group pruned tokens that fall into the same grid cell, as shown in \cref{fig:inference_framework} (b). Within each cell, the pruned tokens are fused into a single token by computing a weighted average of their representations, using their importance scores as weights. This fused token then replaces the pruned token with the highest importance score in that cell and is incorporated into the set of retained tokens.

Formally, a token \(t_i\) with spatial coordinates \((x_i, y_i)\) belongs to a grid cell \(\mathcal{G}_{p,q}\) if

\begin{equation}
    \lfloor x_i / a \rfloor = p \quad \text{and} \quad \lfloor y_i / a \rfloor = q,
\end{equation}
where \(p\) and \(q\) are the row and column indices of the grid cell, respectively. Suppose there are \(k\) pruned tokens in \(\mathcal{G}_{p,q}\) with corresponding importance scores \(I_1, I_2, \dots, I_k\) and token representations \(\mathbf{t}_1, \mathbf{t}_2, \dots, \mathbf{t}_k\). The fused token \(\mathbf{t}_{\text{merged}}^{p,q}\) is computed as:

\begin{equation}
    \mathbf{t}_{\text{merged}}^{p,q}
    = \frac{\sum_{i=1}^{k} I_i \cdot \mathbf{t}_i}
           {\sum_{i=1}^{k} I_i}.
\end{equation}
We then assign \(\mathbf{t}_{\text{merged}}^{p,q}\) to the position of the pruned token with the highest importance score \(I_{\text{max}}\) in \(\mathcal{G}_{p,q}\), counting it among the retained tokens. By combining token pruning with this recycling strategy, we reduce token redundancy while avoiding potential information loss.

\subsection{Pruning Strategy Optimization}
\label{sec:3_3}
The pruning strategies described in \cref{sec:3_1} and \cref{sec:3_2} involve several key hyperparameters that directly affect the performance of LMMs after token pruning. Properly defining these hyperparameters is crucial for preserving model performance. However, existing approaches often rely on manually designed pruning strategies and apply the same strategy across different LMMs, without considering the unique characteristics of each model. This coarse-grained approach can lead to significant performance degradation due to its lack of fitness for different LMMs, leaving the task of designing pruning strategies tailored to specific LMMs as a formidable challenge. 

Previous interpretability studies of LMMs~\cite{kaduri2024s,zhang2024cross} offer crucial insights into their internal mechanisms. Specifically, these works reveal that in the lower and middle layers of LMMs, visual information from visual tokens is aggregated into the corresponding query text tokens. In the higher layers, the multimodal representation encoded in the query text tokens is further progressively propagated to the final position of the input sequence, ultimately influencing the subsequent inference process.

Inspired by this insight, we propose a framework that requires only a small amount of unlabeled data and leverages the internal flow of visual information within LMMs to search for the superior pruning strategy tailored to the characteristics of different LMMs, as shown in \cref{fig:inference_framework} (2). Specifically, we recast the task of designing the superior pruning strategy as an optimization problem, aiming to minimize the discrepancy in visual information flow with and without applying a pruning strategy to perform visual token pruning. In this framework, we treat the final token in the last layer as the representative outcome of the visual information flow. We use cosine similarity to measure the similarity of the final token representation with and without visual token pruning. A higher similarity indicates that the discrepancy between the visual information flow with and without pruning is smaller. Accordingly, we define the following optimization objective:
\begin{equation}
\max_{s \in \mathcal{S}} \  
f(s) = \text{CosineSim}(h_f, g_s(h_f)).
\end{equation}
Here, \(h_f\) represents the representation of the final token in the last layer before pruning, and \(g_s(h_f)\) represents the final token representation after applying the pruning strategy \(s\) (with \(g_s\) modeling its effect on the token feature), and \(\mathcal{S}\) is the solution space.
To efficiently search for the superior pruning strategy \(s\), we employ Bayesian optimization, which systematically explores the hyperparameter space—including the threshold sensitivity \(t\), the entropy weight \(\alpha\), the grid size \(a\), and the pruning ratios \(R_1\), \(R_2\), and \(R_3\)—to maximize the target function \(f(s)\). In short, Bayesian optimization constructs a surrogate model to approximate the objective function and employs an acquisition function to balance exploration and exploitation, thereby efficiently guiding the search for promising hyperparameter settings. The details of this optimization process are presented in \cref{algo:bayesian_optimization}.

%% file: sec/4_experiments.tex
\section{Experiments}
In this section, we evaluate our approach on various LMMs across diverse image and video benchmarks, followed by an efficiency analysis and an ablation study of each component. Finally, we illustrate how our method affects discrepancies in the visual information flow with and without pruning, offering deeper insights into our approach.

\subsection{Experiment Setting}
\textbf{Datasets.} We evaluate our method on ten image-based multimodal benchmarks: GQA~\cite{ainslie2023gqa}, VizWiz~\cite{gurari2018vizwiz}, ScienceQA-IMG~\cite{lu2022learn}, TextVQA~\cite{singh2019towards}, ChartQA~\cite{ChartQA},\begin{algorithm}[t!]
    \caption{VFlowOpt with Bayesian Optimization}
    \label{algo:bayesian_optimization}
    \begin{algorithmic}[1]
        \Require  
        LMM $\theta$,  
        data samples $\mathcal{D}$,  
        number of Bayesian optimization iterations $T$,  
        computation budget $\overline{R}$,  
        target function $f(\cdot)$: the sum of cosine similarities between the last token in the last layer of the LMM $\theta$ with and without visual token pruning, computed over all data samples $\mathcal{D}$.  

        \Ensure retention rates $R_1$, $R_2$, $R_3$; threshold sensitivity $t$; entropy weight in importance score $\alpha$; grid size $a$.

        \State Initialize a Gaussian Process model $\mathcal{GP}$  
        \State Define the acquisition function $A(\cdot)$ (Expected Improvement is adopted)  
        \State Uniformly sample initial points:  
        \[
        X_0 = \{(R_1, R_2, t, \alpha, a) \mid R_v, R_1, t, \alpha, a \in \text{valid ranges}\}
        \]
        \ForAll{$x \in X_0$}
            \State Calculate $R_3$ using the constraint: 
            $\overline{R} = (R_1 \cdot L_1 + R_1 \cdot R_2 \cdot L_2 + R_1 \cdot R_2 \cdot R_3 \cdot L_3) / L$
            \State Form pruning strategy: $s = (R_1, R_2, R_3, t, \alpha, a)$
            \State Evaluate the target function $f(s \mid \theta, \mathcal{D})$
        \EndFor
        \For{$n = 0$ to $T-1$}
            \State Fit $\mathcal{GP}$ to the observed data $(S_n, f(S_n \mid \theta, \mathcal{D}))$
            \State Select the next point: $x_{n+1} \gets \arg\max_x A(x; \mathcal{GP})$
            \State Calculate $R_3$ using the constraint: 
            $\overline{R} = (R_1 \cdot L_1 + R_1 \cdot R_2 \cdot L_2 + R_1 \cdot R_2 \cdot R_3 \cdot L_3) / L$
\State Form pruning strategy: {\scriptsize $s_{n+1} = (R_1, R_2, R_3, t, \alpha, a)_{n+1}$}
            \State Update $S_{n+1} \gets S_n \cup \{s_{n+1}\}$
        \EndFor \\
        \Return $(R_1, R_2, R_3, t, \alpha, a)^*$ that maximize $f$
    \end{algorithmic}
\end{algorithm} POPE~\cite{li2023evaluating}, MME~\cite{fu2024mmecomprehensiveevaluationbenchmark}, MMBench~\cite{liu2024mmbench}, MMStar~\cite{MMstar}, and DocVQA~\cite{DocVQA}. For video understanding, we adopt two datasets—SeedBench (video)~\cite{Seedbench} and VideoMME~\cite{VideoMME}, where VideoMME is partitioned by video length into short, medium, and long subsets. Further details are presented in the Appendix.\\\begin{table*}[t!]
    \centering
    \renewcommand{\arraystretch}{1.2}
    \setlength{\tabcolsep}{3pt}
\scalebox{0.98}{\tablestyle{3.8pt}{0.9}

    \begin{tabularx}{\textwidth}{l|*{8}{>{\centering\arraybackslash}X} >{\centering\arraybackslash}p{1.2cm} >{\centering\arraybackslash}p{1.2cm}|>{\centering\arraybackslash}p{1.0cm}}
        \toprule
        \textbf{Method} & \textbf{MMStar}  & \textbf{MME}  & \textbf{MMB}  & \textbf{SQA}  & \textbf{POPE}  & \textbf{GQA}  & \textbf{VizWiz}  & \textbf{VQA}\textsuperscript{\textnormal{Text}}  & \textbf{ChartQA}  & \textbf{DocVQA}  & \textbf{Avg.}  \\
        \midrule
        \rowcolor{lightgray} \multicolumn{12}{c}{\textit{Upper Bound, 100\% Tokens}} \\
        Vanilla & 61.7 & 1581 & 80.8 & 95.8 & 89.1 & 62.2 & 60.4 & 76.0 & 80.0 & 87.5 & 100\% \\
        \midrule
        \rowcolor{lightgray} \multicolumn{12}{c}{\textit{Retain 50\% Tokens ($\downarrow$ 50\%)}} \\
        FastV\texttt{\scriptsize{(ECCV24)}} & 58.9 & 1549 & 79.4 & 92.8 & 87.9 & 61.5 & \textbf{61.1} & 72.5 & 68.6 & 84.0 & 96.5\%\\
        SparseVLM\texttt{\scriptsize{(2024.10)}} & 59.8 & 1577 & 80.5 & 94.1 & 88.1 & 61.9 & 60.4& 73.9 & 70.5 & 80.8 & 97.1\%\\
        VisionZip\texttt{\scriptsize{(CVPR25)}} & 60.4 & 1587 & 80.3 & 94.6 & 89.3 & \textbf{62.7} & 59.8 & 74.2 & 75.4 & 88.4 & 98.9\%\\
        VFlowOpt & \textbf{61.3} & \textbf{1591} & \textbf{81.1} & \textbf{95.4} & \textbf{89.4} & 62.4 & 60.0 & \textbf{75.1} & \textbf{77.8} & \textbf{90.0} & \textbf{99.9\%}\\
        \midrule
        \rowcolor{lightgray} \multicolumn{12}{c}{\textit{Retain 25\% Tokens ($\downarrow$ 75\%)}} \\
        FastV\texttt{\scriptsize{(ECCV24)}}  & 54.0 & 1539 & 77.0 & 88.6 & 83.8 & 58.2 & \textbf{61.0} & 58.3 & 42.7 & 62.9 & 86.4\%\\
        SparseVLM\texttt{\scriptsize{(2024.10)}} & 56.6 & 1520 & 78.7 & 90.3 & 87.2 & 59.7 & 60.8 & 66.3 & 54.0 & 66.6 & 90.5\%\\
        VisionZip\texttt{\scriptsize{(CVPR25)}} & 54.6 & 1562 & 78.9 & 90.4 & 88.8 & 61.0 & 60.4 & 70.0 & 66.3 & 79.6 & 94.3\%\\
        VFlowOpt & \textbf{57.8} & \textbf{1570} & \textbf{79.9} & \textbf{92.3} & \textbf{89.1} & \textbf{61.2} & 60.4 & \textbf{72.5} & \textbf{69.1} & \textbf{82.3} & \textbf{96.3\%}\\
        \midrule
        \rowcolor{lightgray} \multicolumn{12}{c}{\textit{Retain 10\% Tokens ($\downarrow$ 90\%)}} \\
        FastV\texttt{\scriptsize{(ECCV24)}}  & 46.0 & 1209 & 70.1 &81.7 & 77.0 & 51.5 & 56.4 & 35.6 & 21.3 & 33.2 & 69.7\%\\
        SparseVLM\texttt{\scriptsize{(2024.10)}} & 45.1 & 1191 & 71.8 & 83.7 & 80.0 & 54.6 & 56.4 & 39.8 & 37.6 & 39.6 & 74.0\%\\
        VisionZip\texttt{\scriptsize{(CVPR25)}} & 49.5 & 1389 & 74.8 & 86.2 & \textbf{86.1} & 57.2 & 56.8 & 56.4 & 46.1 & 49.0 & 82.1\%\\
        VFlowOpt & \textbf{52.0} & \textbf{1464} & \textbf{75.1} & \textbf{88.4} & 85.2 & \textbf{57.3} & \textbf{57.3} & \textbf{60.2} & \textbf{53.6} & \textbf{56.1} & \textbf{85.5\%}\\
        \bottomrule
    \end{tabularx}}
    \caption{Performance comparison on LLaVA-OneVision-7B under different token retention conditions. ``Avg.'' refers to average accuracy on 10 benchmarks. For each reduction ratio, the best average performance is shown in \textbf{bold}.}
    \label{tab:main_results}
\end{table*}\textbf{Model Architectures.} We integrate VFlowOpt into multiple LMMs, including LLaVA-OneVision-7B, LLaVA-NeXT-7B~\cite{liu2024llavanext}, and Qwen2-VL-7B~\cite{wang2024qwen2}. These models employ various vision encoders (SigLIP, CLIP, and a ViT designed for Qwen2-VL) and LLM backbones (Qwen2-7B and Vicuna-7B). To prevent out-of-memory issues in Qwen2-VL, we set \textit{max\_pixels = 3000000}.\\
\textbf{Comparison Methods.} We compare VFlowOpt against three baseline methods: FastV~\cite{FastV}, 
SparseVLM~\cite{Sparsevlm}, and VisionZip~\cite{Visionzip}. FastV, 
and SparseVLM rely on text-visual attention in the LLM to prune visual tokens but differ as follows: FastV performs a single pruning step after the second LLM layer; 
and SparseVLM uses the attention weights of preselected text tokens to evaluate the importance of visual tokens. VisionZip, in contrast, determines token importance based on the [cls] token’s attention; for models lacking a [cls] token, we follow VisionZip’s original procedure by computing the average attention each token receives from every other token in the ViTs.\\
\textbf{Implementation Details.} 
For LLaVA-OneVision-7B and Qwen2-VL-7B, we perform token pruning at three distinct points: before the LLM, and after the 9th and 18th layers. For LLaVA-NeXT-7B, pruning is conducted before the LLM, and again after the 10th and 20th layers. During optimization, we sample 30 unlabeled instances from each model’s training datasets; for models without publicly available training dataset, we instead use random samples from the LLaVA-OneVision training set. The optimization is performed for a total of 50 iterations. For LMMs that modify visual tokens output by the vision encoder (e.g., unpadding and interpolation in LLaVA-OneVision, and unpadding in LLaVA-NeXT), we apply corresponding transformations to the importance maps so they remain fully aligned with the final visual tokens. 
\subsection{Image Understanding Tasks}

We evaluate the proposed VFlowOpt on image understanding benchmarks with LLaVA-OneVision-7B using various pruning ratios, and present the results in \cref{tab:main_results}. Compared to other baselines, VFlowOpt consistently maintains superior accuracy across different levels of token pruning. With 50\% token retention, VFlowOpt achieves 99.9\% of the original performance, exceeding the second-best approach by 1.0\%. This negligible performance drop underscores the method’s strong potential for practical deployment. Under more extreme pruning conditions with very few retained tokens, VFlowOpt’s advantage becomes more pronounced. When only 10\% of the visual tokens are retained, VFlowOpt preserves 85.5\% of the original performance, surpassing the second-best approach by 3.3\%. This finding suggests that VFlowOpt can effectively leverage limited visual information to maintain high performance under strict computational budgets. 
In contrast, VFlowOpt demonstrates a clear advantage by tailoring its pruning strategy to each model’s unique characteristics.

To further validate VFlowOpt’s generalization capability, we evaluate it on LLaVA-NeXT-7B and Qwen2-VL-7B. As shown in \cref{tab:llava_next_results} and \cref{tab:qwen2vl_results}, VFlowOpt once again delivers the top results on these LMMs. When retaining 25\% of the tokens, VFlowOpt preserves 98.8\% of the original performance on LLaVA-NeXT-7B and 97.8\% on Qwen2-VL-7B. Even under the stringent condition of retaining only 10\% of the tokens, VFlowOpt maintains 93.4\% of the performance on LLaVA-NeXT-7B and 92.8\% on Qwen2-VL-7B, demonstrating its strong generalizability.

\begin{table}[t!]
    \centering
    \renewcommand{\arraystretch}{1.2}
    \setlength{\tabcolsep}{3pt}
\scalebox{0.94}{\tablestyle{3.0pt}{0.9}

    \begin{tabular}{l|cccccc|c}
        \toprule
        \textbf{Method} & \textbf{MMStar}  & \textbf{MME}  & \textbf{MMB}  & \textbf{SQA}  & \textbf{POPE}  & \textbf{GQA}  & \textbf{Avg.}  \\
        \midrule
        \rowcolor{lightgray} \multicolumn{8}{c}{\textit{Upper Bound, 100\% Tokens}} \\
        Vanilla & 37.6 & 1519 & 67.4 & 70.1 & 86.5 & 64.2 & 100\% \\
        \midrule
        \rowcolor{lightgray} \multicolumn{8}{c}{\textit{Retain 25\% Tokens ($\downarrow$ 75\%)}} \\
        FastV  & 35.1 & 1477 & 65.6 & 67.4 & 83.1 & 60.4 & 95.7\% \\
        VisionZip & 35.8 & 1501 & 65.4 & \textbf{67.9} & 86.7 & 61.5 & 97.3\% \\
        VFlowOpt & \textbf{37.0} & \textbf{1514} & \textbf{67.0} & 67.7 & \textbf{87.6} & \textbf{62.6} & \textbf{98.8\%} \\
        \midrule
        \rowcolor{lightgray} \multicolumn{8}{c}{\textit{Retain 10\% Tokens ($\downarrow$ 90\%)}} \\
        FastV  & 29.2 & 1282 & 61.6 & 63.8 & 71.7 & 55.9 & 85.7\% \\
        VisionZip & 32.6 & 1378 & 61.5 & 67.1 & 83.5 & 57.0 & 91.6\% \\
        VFlowOpt & \textbf{35.1} & \textbf{1393} & \textbf{62.9} & \textbf{67.4} & \textbf{83.6} & \textbf{57.3} & \textbf{93.4\%} \\
        \bottomrule
    \end{tabular}}
    \caption{Comparative experiments on LLaVA-NeXT-7B.}
    \label{tab:llava_next_results}
\end{table}

\begin{table}[t!]
    \centering
    \renewcommand{\arraystretch}{1.2}
    \setlength{\tabcolsep}{3pt}
\scalebox{0.94}{\tablestyle{3.0pt}{0.9}
    \begin{tabular}{l|cccccc|c}
        \toprule
        \textbf{Method} & \textbf{MMStar}  & \textbf{MME}  & \textbf{MMB}  & \textbf{SQA}  & \textbf{POPE}  & \textbf{GQA}  & \textbf{Avg.}  \\
        \midrule
        \rowcolor{lightgray} \multicolumn{8}{c}{\textit{Upper Bound, 100\% Tokens}} \\
        Vanilla & 57.5 & 1680 & 80.3 & 84.7 & 88.4 & 62.2  & 100\%\\
        \midrule
        \rowcolor{lightgray} \multicolumn{8}{c}{\textit{Retain 25\% Tokens ($\downarrow$ 75\%)}} \\
        FastV  & 54.5 & 1597 & 76.3 & 79.0 & 81.8 & 57.2 & 93.8 \\
        VisionZip & 55.2 & 1618 & 78.9 & 81.3 & 86.9 & 60.0 & 96.8 \\
        VFlowOpt & \textbf{55.9} & \textbf{1659} & \textbf{79.8} & \textbf{81.5} & \textbf{87.1} & \textbf{60.4} & \textbf{97.8} \\
        \midrule
        \rowcolor{lightgray} \multicolumn{8}{c}{\textit{Retain 10\% Tokens ($\downarrow$ 90\%)}} \\
        FastV  & 44.9 & 1405 & 70.0 & 75.5 & 75.8 & 51.6 & 84.4 \\
        VisionZip & 49.3 & 1518 & 76.5 & 78.2 & 84.3 & 54.1 & 90.9 \\
        VFlowOpt & \textbf{51.0} & \textbf{1591} & \textbf{78.0} & \textbf{78.6} & \textbf{84.5} & \textbf{54.8} & \textbf{92.8} \\
        \bottomrule
    \end{tabular}}
    \caption{Comparative experiments on Qwen2-VL-7B.}
    \label{tab:qwen2vl_results}
\end{table}
\begin{table}[t!]
    \centering
    \renewcommand{\arraystretch}{1.2}
    \setlength{\tabcolsep}{3pt}
    \scalebox{0.94}{\tablestyle{8pt}{0.9}
    \begin{tabular}{l|c|ccc|c}
        \toprule
        \textbf{Method} & 
        \multicolumn{1}{c|}{\textbf{SeedBench}} & 
        \multicolumn{3}{c|}{\textbf{VideoMME}} & 
        \textbf{Avg.} \\
        \cmidrule(r){2-2} \cmidrule(r){3-5}
        & \textbf{(video)} & \textbf{S} & \textbf{M} & \textbf{L} & \\
        \midrule
        \rowcolor{lightgray} \multicolumn{6}{c}{\textit{Upper Bound, 100\% Tokens}} \\
        Vanilla   & 56.9 & 70.5 & 54.6 & 49.5 &100\%\\
        \midrule
        \rowcolor{lightgray} \multicolumn{6}{c}{\textit{Retain 25\% Tokens ($\downarrow$ 75\%)}} \\
        FastV & 54.7 & 66.9 & 53.2 & 47.7 & 96.2\% \\
        VisionZip & 56.4 & 68.3 & 55.3 & 49.0 & 99.0\% \\
        VFlowOpt  & \textbf{56.8} & \textbf{68.9} & \textbf{55.8} & \textbf{49.2} & \textbf{100\%} \\
        \midrule
        \rowcolor{lightgray} \multicolumn{6}{c}{\textit{Retain 10\% Tokens ($\downarrow$ 90\%)}} \\
        FastV     & 48.7 & 53.7 & 47.6 & 42.3 & 83.6\% \\
        VisionZip & 55.0 & 59.7 & 51.8 & 46.3 & 92.4\% \\
        VFlowOpt  & \textbf{55.5} & \textbf{63.3} & \textbf{52.7} & \textbf{48.6} & \textbf{95.5\%} \\
        \bottomrule
    \end{tabular}}
    \caption{Comparative experiments on video understanding tasks.}
    \label{tab:llava_ov_video_results}
\end{table}

\subsection{Generalization to Video Tasks}
We further explore VFlowOpt’s generalization across different modalities by evaluating it on video benchmarks using LLaVA-OneVision-7B. 
As shown in \cref{tab:llava_ov_video_results}, VFlowOpt surpasses other baselines under various token retention ratios. Notably, when only 25\% of the tokens are preserved, VFlowOpt incurs virtually no performance loss on LLaVA-OneVision-7B, maintaining 100\% of its original performance. Even under the more stringent condition of retaining just $10\%$ of the tokens, VFlowOpt still retains 95.5\% of the original performance, underscoring its robust generalization in the video domain.

\subsection{Efficiency Analysis}
We demonstrate the efficiency of VFlowOpt by conducting a comparative study on LLaVA-OneVision-7B running on a single NVIDIA A100-SXM4-80GB GPU, focusing on FLOPs, KV-Cache memory usage, and inference latency. The efficiency analysis of other baseline methods is provided in the Appendix. 

We assess various token pruning ratios on the MME benchmark, measuring overall performance and average efficiency metrics across all samples. As shown in \cref{tab:inference_costs}, VFlowOpt significantly enhances LLaVA-OneVision-7B’s computational efficiency by reducing FLOPs, shrinking KV-Cache memory usage, and accelerating inference speed. Notably, at a 50\% token pruning ratio, VFlowOpt achieves a 49.5\% reduction in KV-Cache memory and a $1.8\times$ speedup in inference, while boosting performance by 0.7\%. At a more aggressive 75\% pruning ratio, we observe a 74.2\% reduction in KV-Cache memory and a $3.1\times$ speedup, with only a 0.7\% performance drop. Under extremely tight computational budgets, pruning 90\% of visual tokens reduces the KV-Cache memory footprint by 89\%, accelerates inference by $3.8\times$, and degrades performance by only 7.4\%. 

Such reductions in inference latency substantially benefit user-facing applications demanding real-time performance, such as autonomous driving and robotics. Moreover, by significantly shrinking the KV-Cache, large-batch inference on LMMs can accommodate more user requests simultaneously, thereby substantially reducing overall inference costs. In summary, VFlowOpt uses far less GPU memory and delivers faster inference while preserving model performance, offering a highly practical solution for efficiently deploying LMMs in real-world scenarios.
\begin{table*}[t!]
    \centering
\scalebox{0.94}{\tablestyle{3.0pt}{0.9}
    \begin{tabular}{@{}l|c|cc|cc|cc|cc@{}}
        \toprule
        \textbf{Methods} & \textbf{Token} & \textbf{FLOPs ↓} & $\Delta$ & \textbf{Latency ↓} & $\Delta$ & \textbf{KV Cache ↓} & $\Delta$ & \textbf{Performance ↑} & $\Delta$ \\ 
                         &  \textbf{Reduction}  & \textbf{(T)}     &          & \textbf{(ms)}      &          & \textbf{(MB)}       &          &                        &          \\ 
        \midrule
        LLaVA-OneVision-7B &    -      &  71.4   &    -     & 1040.1      &  -      & 1786.4  &     -   & 1581                  &     -  \\ 
        \midrule
        + VFlowOpt         & 50\%      & 37.2  &  \hgreen{-48.0\%}   & 584.2 & \hgreen{-43.8\%}         & 902.8        &  \hgreen{-49.5\%}    & 1591                  &   +0.6\%     \\ 
                           & 75\%      & 19.1  &   \hgreen{-73.2\%}  & 328.5      & \hgreen{-68.4\%}   & 460.6       &  \hgreen{-74.2\%} & 1570             &  -0.7\%  \\ 
                           & 90\%      &  7.7  &   \hgreen{-89.2\%}  & 272.1  &  \hgreen{-73.8\%} & 197.1    & \hgreen{-89.0\%}      & 1464        &  -7.4\% \\ 
        \bottomrule
    \end{tabular}%
    }
    \caption{Efficiency analysis of LLaVA-OneVision-7B with VFlowOpt. The detailed metric includes computation (FLOPs), latency, and KV-Cache memory. ($\Delta$) denotes the reduction ratio.
 .}
    \label{tab:inference_costs}
\end{table*}
\subsection{Ablation Study}
To confirm the contributions of each component in our token pruning strategy, we conduct ablation experiments on LLaVA-OneVision-7B under a computational budget of retaining 25\% of tokens. As shown in \cref{tab:ablation}, removing importance calibration (i.e., directly using the mean attention over all tokens received by each visual token as its importance score), omitting token recycling, or discarding progressive pruning (i.e., maintaining the same number of tokens at each layer) leads to noticeable performance degradation across MMStar, MMBench, and SQA. By contrast, the complete VFlowOpt method consistently exhibits minimal performance loss, highlighting the effectiveness of all its components in pruning redundant tokens while preserving essential visual information.

\begin{figure}[t!] 
  \centering
  \includegraphics[width=0.99\linewidth]{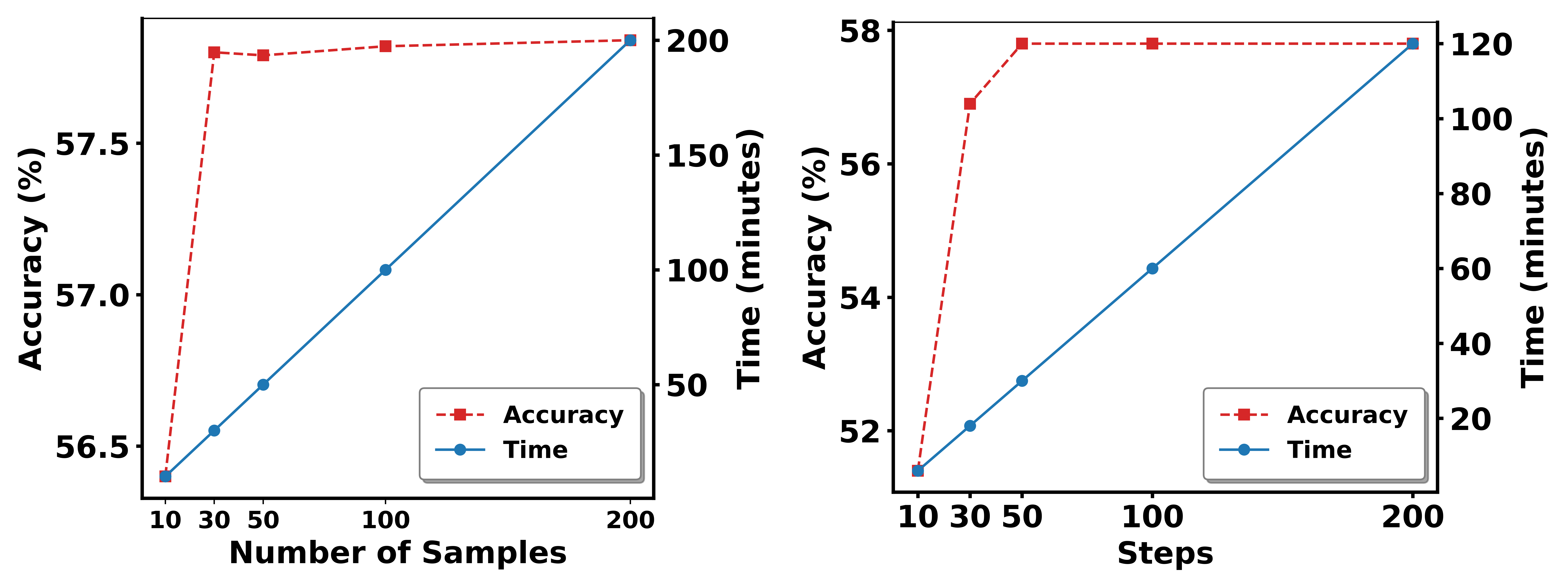} 
  \caption{Relationships of sample size (left) and optimization steps (right) with optimization time and final performance.}
  \label{fig:data_scale_step} %
\end{figure}

We further examine how the number of samples and the number of optimization iterations in the Bayesian optimization procedure influence both optimization time and final performance. As illustrated in \cref{fig:data_scale_step}, the optimization time increases linearly with the number of samples and steps. Using 30 samples and 50 steps strikes an effective balance between time efficiency and performance, requiring only about 30 minutes to reach an exceptional result. More ablation results are provided in the supplementary material.
\begin{table}[t!]
\centering
\scalebox{0.94}{\tablestyle{3.0pt}{0.9}
\begin{tabular}{l|c|c|c}
\toprule
\textbf{Pruning Strategy} & \textbf{MMStar} & \textbf{MMBench} & \textbf{SQA} \\
\midrule 
VFlowOpt              & \textbf{57.8}            & \textbf{79.9}         & \textbf{92.3}      \\ 
\midrule
 w/o Importance Calibration     & 56.2 & 79.4 & 91.8 \\
 w/o Token Merging    & 57.6 & 79.8 & 91.9 \\
 w/o Progressive Pruning & 56.0 & 79.1 & 91.0 \\
\bottomrule
\end{tabular}}
\caption{Ablation studies of pruning strategy components.}
  \label{tab:ablation}
\end{table}

\subsection{Visual Information Flow Analysis}
\begin{figure}[t] 
  \centering
  \includegraphics[width=0.99\linewidth]{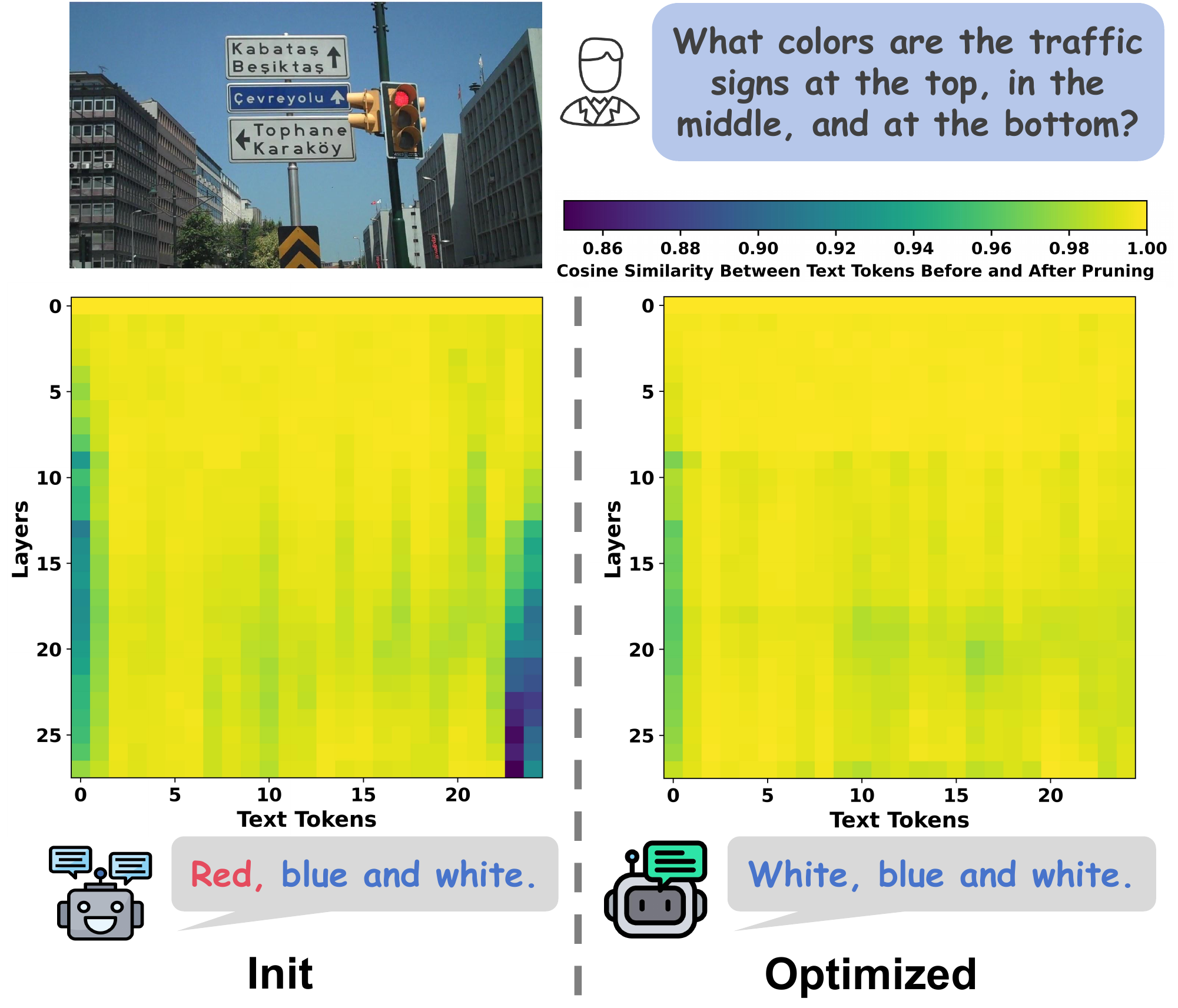} 
  \vspace{-0.3cm}
  \caption{Visualization of the discrepancies in visual information flow corresponding to the pruning strategies before and after optimization, along with an example of visual question answering.}
  \vspace{-0.4cm}
  \label{fig:vflow_visualization} %
\end{figure}

We further investigate how Bayesian optimization formulates the superior pruning strategy by examining its impact on visual information flow. Specifically, we compare the randomly initialized pruning strategy derived from the Bayesian optimization process with its final optimized strategy, measuring their respective effects on the text tokens that follow the visual tokens at each LLM layer. Because these text tokens can receive information from the visual tokens via the LLM’s unidirectional attention mechanism, examining differences in these tokens with and without pruning effectively reveals how visual information flow is altered. As shown in \cref{fig:vflow_visualization}, under the same computational budget, the initial strategy causes a pronounced discrepancy between text token representations with and without pruning, indicating that it significantly disrupts the visual information flow. In contrast, when using the optimized strategy, the text tokens remain much more similar to those observed without visual token pruning. This important finding suggests that the optimization process yields a pruning strategy that preserves crucial visual information by minimizing the difference in visual information flow with and without pruning, thereby retaining LMM performance to the greatest extent possible.




%% file: sec/5_conclusion.tex
\section{Conclusion}
In this paper, we address the pressing challenge of computational inefficiency in LMMs by introducing a general framework for visual token pruning. We reformulate pruning as an optimization problem focused on minimizing the divergence of visual information flow with and without pruning, thereby facilitating a tailored pruning strategy for diverse LMMs. By leveraging calibrated attention in ViTs to evaluate token significance and merging pruned tokens using importance-based weighting, we preserve critical information during inference without requiring retraining or manual tuning. Extensive experiments underscore the efficiency and generalizability of our approach, making it a practical solution for deploying LMMs in real-world settings, particularly in resource-constrained environments.
\clearpage

%% file: sec/suppl.tex
\clearpage
\appendix
\setcounter{page}{1}
\maketitlesupplementary

\section{Overview of Baselines}
\label{sec:dsgd_baselines}
\begin{itemize}

\item FastV\citep{FastV} is a plug-and-play method that optimizes inference efficiency in LMMs by dynamically pruning visual tokens after the second layer, significantly reducing computational costs while maintaining performance. It identifies that image tokens receive drastically lower attention in LLM and strategically removes less impactful tokens.
    \item FitPrune~\citep{Fitprune} is a training-free method for pruning visual tokens in multimodal LMMs, based on quickly estimating optimal pruning schemes through attention distribution fitting. It statistically determines which tokens can be discarded by minimizing divergence between attention distributions before and after pruning, using only a small batch of inference data. This approach rapidly produces a pruning recipe tailored to a given computation budget, significantly reducing computational complexity while preserving model performance.
    \item Pdrop~\citep{Pdrop} accelerates large vision-language models by progressively removing redundant visual tokens in deeper layers based on token similarity. It partitions models into multiple stages, maintaining all tokens initially to preserve critical visual information, then gradually pruning tokens as layers deepen. This approach effectively reduces computational costs without compromising performance during both training and inference.
    \item Sparsevlm~\citep{Sparsevlm} introduces a training-free, text-guided visual token sparsification method for LMMs, significantly reducing computational overhead by adaptively selecting important visual tokens based on relevant text prompts. It employs an adaptive pruning strategy at each layer and recycles pruned visual tokens into compact representations to minimize information loss. 
    \item Visionzip~\citep{Visionzip}  is a simple yet effective method that reduces visual token redundancy in LMMs by selecting only the most informative tokens, significantly improving efficiency while maintaining performance. It employs a text-agnostic approach that merges and compresses redundant tokens, reducing computational costs and enhancing inference speed without requiring additional training.

\end{itemize}

\section{Overview of Benchmarks}
\label{sec:benchmarks}
\begin{itemize}

\item{MME}~\cite{fu2024mmecomprehensiveevaluationbenchmark} offers a robust benchmark for evaluating LVLMs across multimodal tasks. It assesses models on two major fronts: perception and cognition, using 14 well-structured subtasks that challenge their interpretive and analytical abilities.

\item{MMBench}~\cite{liu2024mmbench} takes a two-pronged approach by introducing an extensive dataset that broadens the scope of evaluation questions and a novel CircularEval strategy that utilizes ChatGPT to convert free-form responses into structured answer choices.

\item{ScienceQA}~\cite{lu2022learn} focuses on evaluating multi-hop reasoning and interpretability within scientific domains. It features a large dataset of approximately 21K multiple-choice questions across a variety of science topics, accompanied by detailed annotations and explanations.

\item{VizWiz}~\cite{gurari2018vizwiz} stands out in the VQA field by using a dataset of over 31,000 visual questions that come from a real-world setting, featuring images taken by visually impaired individuals and their associated spoken queries, along with crowdsourced answers.

\item{GQA}~\cite{ainslie2023gqa} is built for complex visual reasoning tasks, containing 22 million questions generated from scene graph-based structures. It incorporates innovative evaluation metrics focused on consistency, grounding, and plausibility, pushing the boundaries of vision-language evaluation.

\item{POPE}~\cite{li2023evaluating} introduces a methodology to evaluate object hallucination in LVLMs, transforming the task into a binary classification problem. By using simple Yes-or-No prompts, POPE highlights model tendencies towards hallucination through various object sampling strategies.

\item{VQA} ~\citep{VQA} collects complementary images such that every question in the balanced dataset is associated with a pair of similar images that result in two different answers to the question. 

\item{ChartQA}~\citep{ChartQA} is a large-scale benchmark designed for question answering on charts, focusing on both visual and logical reasoning with 9.6K human-written and 23.1K automatically generated questions.

\item{DocVQA}~\citep{DocVQA} is a large-scale dataset designed for Visual Question Answering (VQA) on document images, containing 50,000 questions over 12,000+ real-world documents. Unlike previous datasets, it requires models to understand both textual content and visual layout, including tables, forms, and complex structures.
\begin{table*}[h!]
    \centering
\scalebox{0.94}{\tablestyle{3.0pt}{0.9}
    \begin{tabular}{@{}l|c|cc|cc|cc|cc@{}}
        \toprule
        \textbf{Methods} & \textbf{Token} & \textbf{FLOPs ↓} & $\Delta$ & \textbf{Latency ↓} & $\Delta$ & \textbf{KV Cache ↓} & $\Delta$ & \textbf{Performance ↑} & $\Delta$ \\ 
                         &  \textbf{Reduction}  & \textbf{(T)}     &          & \textbf{(ms)}      &          & \textbf{(MB)}       &          &                        &          \\ 
        \midrule
        LLaVA-OneVision-7B &    -      &  71.4   &    -     & 1040.1      &  -      & 1786.4  &     -   & 1581                  &     -  \\ 
        \midrule
        + VFlowOpt         & 50\%      & 37.2  &  \hgreen{-48.0\%}  & 584.2 & \hgreen{-43.8\%}         & 902.8        &  \hgreen{-49.5\%}    & 1591                  &   +0.6\%     \\ 
        + FastV       & 50\%      & 38.1  &   \hgreen{-46.6\%}  & 615.1      & \hgreen{-41.9\%}   & 902.8        &  \hgreen{-49.5\%} & 1549             &  -2.0\%  \\ 
        + VisionZip    & 50\%      &  37.7  &   \hgreen{-47.2\%}  & 580.7  &  \hgreen{-44,2\%} & 902.8        &  \hgreen{-49.5\%}      & 1587        &  +0.1\% \\ 
        \bottomrule
    \end{tabular}%
    }
    \caption{Efficiency analysis of LLaVA-OneVision-7B with VFlowOpt, FastV, and VisionZip. The detailed metric includes computation (FLOPs), latency, and KV-Cache memory. ($\Delta$) denotes the reduction ratio.}
    \label{tab:efficiency_baseline}
\end{table*}
\begin{table*}[ht]
\centering
\scalebox{0.95}{
\begin{tabular}{lcccccccc}
\toprule
 & MMStar & MME & MMB & SQA & POPE & GQA & DocVQA & VQA\textsuperscript{\textnormal{Text}} \\
\midrule
VisionZip & 54.6 & 1562 & 78.9 &  90.4 & 88.8 & 61.0 & 79.6 & 70.0 \\
Ours (Random) & 57.8 & 1570 & 79.9 & 92.3 & 89.1 & 61.2  & 82.3 & 72.5 \\
Ours (\tiny MathV360K-GEOS\normalsize)  & 57.8 & 1566 & 79.8 & 92.0 & 89.1 & 61.0 & 82.1 & 72.8 \\
\bottomrule
\end{tabular}}
\vspace{-1pt}
\caption{Impact of optimization data selection}
\label{tab:impact_optimization_data}
\end{table*}
\item{MMstar}~\citep{MMstar} is a new benchmark designed to address issues in evaluating Large Vision-Language Models (LVLMs), specifically unnecessary visual content and unintentional data leakage, which can mislead performance assessments. It includes 1,500 carefully selected vision-dependent samples, ensuring accurate evaluation of LVLMs' true multi-modal reasoning abilities. MMStar introduces new metrics—Multi-Modal Gain (MG) and Multi-Modal Leakage (ML)—to measure actual improvements from multi-modal training, with evaluations showing GPT-4V leading in both accuracy and multi-modal efficiency.

\item{SeedBench}~\citep{Seedbench} is a large-scale benchmark designed to evaluate the generative comprehension capabilities of Multimodal Large Language Models (MLLMs), featuring 19K human-annotated multiple-choice questions across 12 evaluation dimensions for both images and videos. 
\item{VideoMME}~\citep{VideoMME} is the first comprehensive benchmark designed to evaluate Multi-Modal Large Language Models (MLLMs) in video analysis, covering 900 manually annotated videos across six diverse domains and 30 subcategories. It introduces a full-spectrum evaluation with multi-modal inputs, including subtitles and audio, and assesses models across various temporal contexts, from short clips to hour-long videos. 
\end{itemize}


\section{Efficiency Analysis about Baselines}
We evaluate VFlowOpt, the well-performing baseline FastV, and VisionZip on efficiency metrics under the condition of retaining 50\% of the tokens. With the same token retention rate, all methods showed identical KV-Cache memory usage, while FLOPs and latency exhibited slight differences, as shown in \cref{tab:efficiency_baseline}.

\section{More ablation studies}
\subsection{Choice of the optimization target}
\begin{table}
\centering
\scalebox{0.80}{
\begin{tabular}{lcccccc}
\toprule
 & MMStar & MME & MMB & SQA & POPE & GQA \\
\midrule
\textbf{Last Token} & \textbf{57.8} & \textbf{1570} & \textbf{79.9} & \textbf{92.3} & \textbf{89.1} & \textbf{61.2} \\
Mean Pooling & 56.1 & 1549 & 77.5 & 92.1 & 88.5 & 60.6 \\
First Token  & 54.2 & 1530 & 77.7 & 89.5 & 85.4 & 60.4 \\
Top-3 Tokens & 56.8 & 1544 & 78.6 & 92.3 & 88.3 & 61.1 \\
\bottomrule
\end{tabular}}
\caption{Analysis of choice of the optimization target}
\label{tab:choice_optimization_target}
\end{table}
We are inspired by previous interpretability studies (Main Paper L273–L281) and consider the last token as the most representative one of such interactions. Results (shown in~\cref{tab:choice_optimization_target}) show that optimizing for the last token yields the best performance. We will add this in the revised paper.
\begin{table}[t!]
\centering
\small
\scalebox{0.87}{
\begin{tabular}{lccc}
\toprule
 & DocVQA & VQA\textsuperscript{\textnormal{Text}} & POPE \\
\midrule
VFlowOpt & 82.3 & 72.5 & 89.1 \\
w/o Importance Calibration  & 80.3 & 71.4 & 88.6 \\
w/o Token Merging & 82.0 & 72.4 & 86.8 \\
w/o Progressive Pruning & 81.9 & 71.6 & 88.2 \\
\bottomrule
\end{tabular}}
\vspace{-4pt}
\caption{Ablation studies on more benchmarks}
\label{tab:Ablation_more_benchmarks}
\end{table}

\subsection{Impact of optimization data selection}
The result of our optimization is independent of data selection because the visual information flow being optimized is task-agnostic and model-specific. In our experiments, repeated random sampling yields nearly identical results. To further validate this, we optimize using 30 samples from the task-specific split (MathV360K-Geometry3K) of the LLaVA-OV training data. The model consistently achieves strong results across various tasks, regardless of data selection (shown in~\cref{tab:impact_optimization_data}).

\subsection{Ablation studies on more benchmarks}
Additional results in~\cref{tab:Ablation_more_benchmarks} show that Token Merging is crucial for preserving coarse-grained semantics, while Importance Calibration and Progressive Pruning help maintain fine-grained visual perception.